\documentclass[11pt]{article}

\usepackage[preprint]{acl}
 
\usepackage{times}
\usepackage{latexsym}

\usepackage[T1]{fontenc}

\usepackage[utf8]{inputenc}

\usepackage{microtype}

\usepackage{inconsolata}

\usepackage{graphicx}

\usepackage{tipa}
\usepackage{adjustbox}
\usepackage{array}
\usepackage{tabularx}
\usepackage{algorithm}
\usepackage[noend]{algpseudocode}
\usepackage{amsmath}
\usepackage{booktabs}

\usepackage{xcolor}

\newcolumntype{R}[2]{
    >{\adjustbox{angle=#1,lap=\width-(#2)}\bgroup}
    l
    <{\egroup}
}

\newcolumntype{X}{>{\centering\arraybackslash}X}

\title{Neural Induction of Finite-State Transducers}

\author{Michael Ginn${}^{1}$ \and Alexis Palmer${}^{1}$ \and Mans Hulden${}^{2}$  \\
    ${}^{1}$University of Colorado \quad ${}^{2}$New College of Florida \\
  \texttt{michael.ginn@colorado.edu} \\}

\begin{document}
\maketitle
\begin{abstract}
Finite-State Transducers (FSTs) are effective models for string-to-string rewriting tasks, often providing the efficiency necessary for high-performance applications, but constructing transducers by hand is difficult. In this work, we propose a \textbf{novel method for automatically constructing unweighted FSTs} following the hidden state geometry learned by a recurrent neural network. We evaluate our methods on real-world datasets for morphological inflection, grapheme-to-phoneme prediction, and historical normalization, showing that the constructed FSTs are highly accurate and robust for many datasets, \textbf{substantially outperforming classical transducer learning algorithms by up to 87\% accuracy on held-out test sets}.
\end{abstract}

\section{Introduction}

\textit{Finite-State Transducers (FSTs)} are effective models for string-to-string tasks such as spellchecking and autocomplete \citep{ouyang_mobile_2017}, morphological inflection \citep{golob2012fst}, grapheme-to-phoneme conversion (G2P, \citealp{manohar2022speech}), and transliteration \citep{hellsten_transliterated_2017}. Neural networks have largely supplanted FSTs in research settings, thanks to their improved accuracy, tolerance for noise, and ease of construction from data (e.g. \citealp{kann-schutze-2016-med}). However, FSTs are typically more efficient and less resource-intensive than neural networks, making them an appropriate choice for high-performance applications such as mobile keyboards \citep{wolf-sonkin-etal-2019-latin} and embedded systems \citep{wang2012design}.

\begin{figure}[t]
    \centering
    \includegraphics[width=\linewidth]{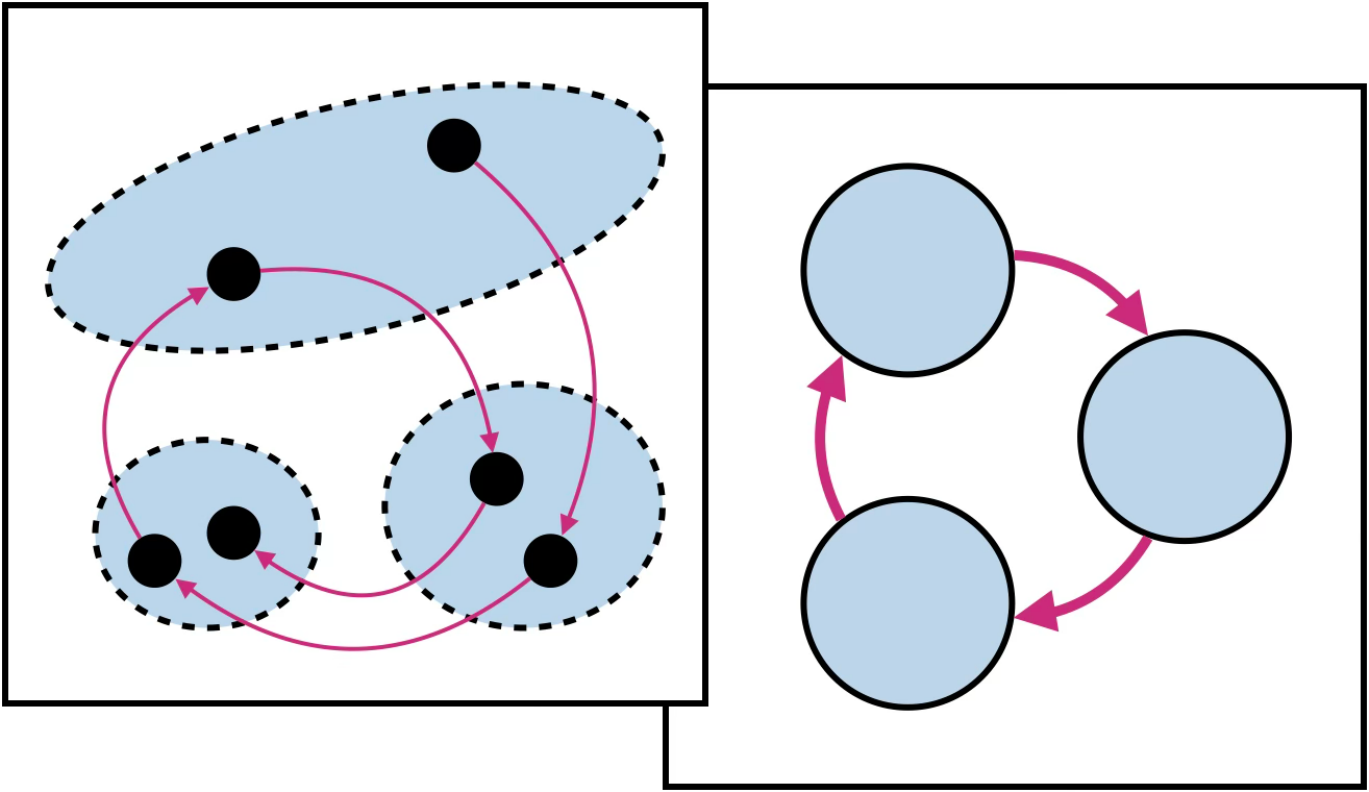}
    \caption{There is a theoretical and empirical correspondence between recurrent neural network's continuous state space (left) and the finite-state space of an automaton (right). In the RNN state space, individual activation values may form clusters that correspond to finite states in the FSA (blue). Transitions between individual activations can be aggregated to form the transitions of the FSA (purple).}
    \label{fig:diagram}
\end{figure}

Construction of FSTs is difficult, requiring domain knowledge and significant human effort---for example, \citet{beemer-etal-2020-linguist} observed that it took approximately forty hours for an expert to hand-craft a single transducer for basic morphological inflection. Prior research has proposed grammar induction algorithms to automatically learn transducers from datasets with pairs of inputs and outputs, but these algorithms generally struggle to generalize robustly \citep{gildea-jurafsky-1995-automatic} or require an oracle to verify arbitrary examples \citep{vilar1996query}.

In this work, we propose a novel approach for \textbf{unweighted transducer induction}\footnote{Weighted transducer induction is a distinct, and typically easier problem, and has a number of solutions discussed in \S \ref{sec:related_work}} which utilizes neural networks as intermediate learners. Neural networks are highly effective at learning from data, even in the presence of noise. Furthermore, there is a clear theoretical correspondence between the continuous, vector-valued hidden state space of a recurrent neural network (RNN) and the discrete state space of a transducer. Our method is inspired by the seminal work of \citet{giles_learning_1992, omlin_extraction_1996}, which sought to understand RNN behavior by forming finite-state automata (FSAs) from clusters of hidden state vectors. Our work is novel in three key ways
\begin{enumerate}
    \item We use real-world, noisy datasets instead of small toy grammars.
    \item We extract FSTs instead of FSAs.
    \item We propose architectural modifications for the intermediate neural networks that facilitate transducer extraction. Specifically, we design a custom transduction training objective and use a spectral penalty to encourage the RNN to model finite-state-esque dynamics.
\end{enumerate}

We evaluate our method on real-world datasets for morphological inflection, grapheme-to-phoneme conversion, and historical normalization. \textbf{We find that our method is highly accurate on inflection}, but struggles on grapheme-to-phoneme and normalization data due to bidirectional dependencies. Nonetheless, our method produces more accurate transducers than existing methods for nearly every dataset in the study. For inflection, the extracted transducers often achieve \textbf{similar accuracy to transducers hand-crafted by human experts}. Our code will be released on GitHub.\footnote{\url{https://github.com/michaelpginn/fst-distillation}}


\section{Background}
\label{sec:background}
A finite-state transducer is defined as a tuple $A=(\Sigma, \Pi, Q, q_0, \delta)$, where:
\begin{itemize}
    \item $\Sigma$ is the \textit{input alphabet} of the automaton, a finite non-empty set of input symbols.
    \item $\Pi$ is the \textit{output alphabet} of the automaton (often the same as the input alphabet).
    \item $Q$ is a finite non-empty set of states.
    \item $q_0 \in Q$ is the \textit{initial state}.
    \item $\delta: Q \times (\Sigma \cup \{\epsilon\}) \rightarrow Q \times (\Pi \cup \{\epsilon\})$ is the \textit{transition function}, which maps a given state and input symbol (or the empty string) to a new state and output symbol (or empty string).
\end{itemize}

\begin{figure}[h]
    \centering
    \includegraphics[width=\linewidth]{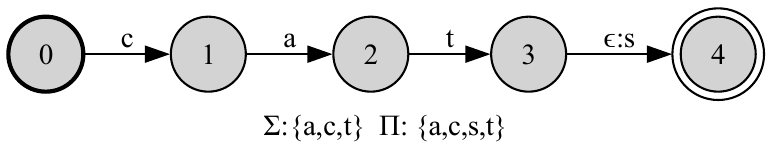}
    \caption{Example FST that rewrites "cat" to "cats".}
    \label{fig:cats}
\end{figure}

\noindent In this work, we focus on \textbf{regular transducers}, which are deterministic/sequential, i.e. a given input string only has one possible output. An example FST is given in \autoref{fig:cats}, which inflects the word `cat' to the plural `cats'. For this FST, $\Sigma = \{c, a, t\}$, $\Pi = \{c, a, t, s\}$, $Q = {0, 1, 2, 3, 4}$, and $q_0 = 0$.

Transducers are closely related to \textit{finite-state automata (FSAs)}, which recognize a set of strings. Automata have a single alphabet of symbols, and each transition is associated with a single symbol rather than an input and output symbol. Every automaton can be formulated as a transducer that performs the identity function, mapping each allowed string to itself. Likewise, every transducer can be formulated as an automaton that recognizes a valid sequence of (input symbol : output symbol) tuples. We make use of this equivalence in order to extend existing automaton induction methods \citep{giles_higher_1989} to transducers.

\section{Method}
\label{sec:method1}

\subsection{Motivation}

Early work on recurrent neural networks (RNNs) recognized an intuitive correspondence between RNNs and finite-state automata \citep{cleeremans1989finite, giles_higher_1989, pollack_induction_1991}. Both models process input sequences symbol-by-symbol, moving through a state space (continuous for RNNs and discrete for FSAs) according to a transition function. These similarities motivated attempts to extract automata from RNNs using a method known as \textit{state clustering}. The method, first proposed by \citet{giles_extracting_1991, giles_learning_1992}, clusters hidden-state values to form the states of the automaton (\autoref{fig:diagram}). These attempts were highly successful at recovering automata from RNNs trained on synthetic data generated by regular languages such as the Tomita grammars \citep{tomita_learning_1982}. Furthermore, \citet{casey_dynamics_1996} proved that an RNN which robustly learns to model a regular language \textit{must} organize its state space to correspond with the target FSA.

\subsection{Overview}
We propose an algorithm for 
FST inference inspired by these earlier 
findings. Prior work focused on interpretability, using state clustering to understand the internal dynamics of standard RNN models. Because we are more interested in creating high-accuracy transducers, 
we modify the architecture and training process of the RNN to facilitate transducer extraction. Our method is as follows:

\begin{enumerate}
    \item Perform \textbf{alignment} of input and output strings (\S \ref{sec:alignment_step}).
    \item Train an RNN on \textbf{transduction} (\S \ref{sec:train_rnn_step}).
    \item Collect \textbf{hidden-state values} from the RNN for input strings from the training set and additional synthetically-generated input strings.
    \item Perform \textbf{clustering} on hidden-state values to form the states of the FST (\S \ref{sec:clustering}).
    \item For each cluster, select the most common transition per input symbol, and use a \textbf{state splitting algorithm} to resolve non-deterministic transitions (\S \ref{sec:splitting}).
\end{enumerate}

\subsection{Steps}
\subsubsection{Alignment}
\label{sec:alignment_step}
A unique challenge in transducer induction 
is that an \textit{alignment} between pairs of input and output characters must be learned in addition to learning the structure of the automaton. In our method, alignment learning is performed prior to structure learning.
We perform alignment of all the input and output strings in the \textbf{training dataset only} using the \textit{Chinese Restaurant Process Alignment (CRPAlign)}, a Bayesian Monte Carlo alignment algorithm first described in \citet{cotterell-etal-2016-sigmorphon}. We provide this algorithm in \autoref{sec:alignment}. 

For examples where the input and output string are the same length, the alignment algorithm simply aligns each pair of characters. For example, given \texttt{(run, ran)}, the alignment would be:
\begin{center}
    \texttt{r u n} \\
    \texttt{r a n}
\end{center}
\noindent However, if the input and output strings are different lengths, the method will insert \textit{epsilons ($\epsilon$)}, which represent an empty string. For \texttt{(run, runs)}, the alignment might be:
\begin{center}
    \texttt{r u n $\epsilon$} \\
    \texttt{r u n s}
\end{center}
Epsilons in the output string are left alone. Epsilons in the input string, though, are problematic
as they can introduce non-determinism when we later convert the aligned pairs into transducer transitions.\footnote{Because an epsilon-input transition is optional, potentially resulting in multiple possible paths through the automaton if there is also a valid non-epsilon transition from a given state.} To solve this, we remove epsilons in the input string and merge the corresponding outputs. We use two merging strategies. For G2P and normalization, we always merge outputs to the right:
\begin{center}
\begin{tabular}{r c l}
\texttt{a $\epsilon$ c} & $\rightarrow$ & \texttt{a c} \\
\texttt{x y z}          & $\rightarrow$ & \texttt{x yz}
\end{tabular}
\end{center}
For inflection, we use a greedy merging strategy based on global probabilities. We count all pairs of aligned characters involving epsilon inputs, merge the most common pair, and repeat until no more epsilon inputs remain. In the preceding example, if \texttt{(a,x)($\epsilon$,y)} is the most common pair, the merged example would be:
\begin{center}
\begin{tabular}{r c l}
\texttt{a $\epsilon$ c} & $\rightarrow$ & \texttt{a~ c} \\
\texttt{x y z}          & $\rightarrow$ & \texttt{xy z}
\end{tabular}
\end{center}



These aligned pairs form the transitions of the FST. For example, an aligned pair \texttt{(a,xy)}  would correspond to a transition with
input \texttt{a} and output \texttt{xy}. Next, we will use an RNN to induce the structure of the transducer, given these alignments.

\subsubsection{RNN Training}
\label{sec:train_rnn_step}









We use one-layer simple RNNs (also known as \textit{Elman RNNs}). RNNs can be trained on different tasks/training objectives, with implications for the types of hidden states that are learned. Virtually all prior work trained RNNs on binary classification, where the RNN should predict whether or not a string is a valid member of the target formal language \citep{giles_extracting_1991, giles_learning_1992, watrous_induction_1992, omlin_extraction_1996, weiss_extracting_2018}. This approach works well on simple datasets, but we hypothesized, and then empirically verified, that it 
does not induce well-separated hidden states for challenging, real-world datasets.

Instead, we propose a novel \textbf{transduction training objective} that directly matches the structure of a finite-state transducer, providing explicit supervision of every hidden state in a training example. We formulate the transduction task as follows. For each position in the aligned sequence, we compute the hidden state using the standard update rule.
\begin{equation}
    h_t = \sigma (W_hh_{t-1} + W_xx_t)
\end{equation}
\noindent where $h_t$ is the hidden state at time $t$, $x_t$ is the input symbol embedding, and $W_h, W_x$ are trainable weight matrices. At each timestep, we train the model to predict the next output symbol given some input symbol. We use a linear layer that takes a concatenated hidden state and the embedding for the \textbf{next input symbol} $x_{t+1}$ and predicts the corresponding output symbol $y_{t+1}$:
\begin{equation}
    p(y) = \textnormal{softmax}(W_y \cdot \textnormal{concat}(h_{t}, x_{t+1}) )
\end{equation}
Our objective function is standard cross-entropy loss over the output symbol logits, $CEL(p(y), y^*)$ where $y^*$ is the true output symbol. This training objective closely matches key features of an FST: intermediate states are solely determined by the sequence of input symbols up to that point, and states are differentiated by their outgoing transitions. In addition, we use a \textbf{spectral norm} penalty $\mathcal{L}_{SN}$, estimated using the power iteration on the weights of each linear layer for updating the hidden state \citep{miyato2018spectral}. We hypothesized that by encouraging a small spectral norm, and thus a small Lipschitz constant for the hidden state update function, the RNN will tend to have finite-state-like dynamics (i.e. fixed point attractors), and we found empirically that this enabled more robust FST extraction. We weight the spectral norm penalty using a hyperparameter $\lambda_{SN}$, giving our full objective as:
\begin{equation}
\mathcal{L} = \mathcal{L}_{CE}(p(y), y^*) + \lambda_{SN} \mathcal{L}_{SN} 
\end{equation}

During training, our token vocabulary includes any merged symbols produced during alignment, unmerged single-character symbols, and a token for each discrete morphological tag. We do not perform any additional tokenization or merging.

We choose to use Elman RNNs because their simple update rule imposes a Markov assumption, where the next hidden state is dependent only on the current state. Gated RNNs (LSTMs and GRUs) violate this assumption, since by design, the memory cells retain information about the full sequence of prior states. Furthermore, \citet{wang_comparative_2018} find that 
these complex hidden-state dynamics make extracting finite-state automata more difficult. We find empirically that increasing the number of layers does not improve the extracted FST. Details for the training process in \autoref{sec:rnn_training_appendix}.



\subsubsection{Collecting Hidden States}
\label{sec:hidden_states_step}

Next, we collect hidden state activation values obtained by inputting strings to the RNN. We use two sources of input-side strings: 1) the examples in the training dataset, and 2) synthetically-generated strings that are plausible for the task. 

The latter is necessary because of a difference in how neural models and FSTs encode their \textit{domain}, i.e. the strings that can be inputted to the model. An RNN can produce an output for any input string, allowing it to generalize to strings outside of its training set. Meanwhile, an FST must explicitly encode the set of accepted strings through the input side of the transition labels. If a string does not have a valid sequence of transitions, the FST cannot produce an output. Thus, if we only collect activations for the strings in the training set, 
we are very unlikely to observe every possible transition.
To solve this, we create additional strings that provide greater coverage over the domain.

The particular strategy we use for synthetic string generation depends on the task.
For inflection, the domain is finite, consisting of a fixed number of lemma and feature tag combinations, so we can generate synthetic strings by swapping feature tags for lemmas in the training set. For G2P and normalization, the domain is theoretically infinite. We approximate the domain by creating an n-gram language model from the training set strings. Then, we generate all possible strings from these n-grams up to a max length (we use 6 for efficiency).

Finally, for every training and synthetic example, we feed the string to the RNN and collect the hidden state activation values and predicted output string (ignoring predictions for the training set, where we have a gold output already). Every symbol in the input string is assigned a hidden state, which is computed by recursively applying the update rule for all of the preceding symbols.

\subsubsection{State Clustering}
\label{sec:clustering}

Thanks to the training objective, we expect hidden states that are close together to be more likely to correspond to the same state of the transducer. We standardize activations and perform k-means clustering \citep{mcqueen1967some}. We also tested other clustering algorithms (OPTICS, DBSCAN) but found no clear performance benefit. 

The clusters of activations form the states of the FST. Next, we create transitions between states by aggregating the transitions between activations assigned to each cluster (\autoref{fig:splitting}, part 1). For the activations collected from the training data, we use the ground-truth input and output labels. For the activations predicted for the full domain, we use the output labels as predicted by the RNN transducer. 

A key desideratum for our extracted transducer is \textbf{input determinism}. In some cases, activations within a single cluster may violate input determinism and have multiple transitions for the same input symbol with different outputs or different destination states. In this case, we try to recover using the state splitting algorithm described in the following section.

\subsubsection{State Splitting}
\label{sec:splitting}

\begin{figure}[t]
    \centering
    \includegraphics[width=\linewidth]{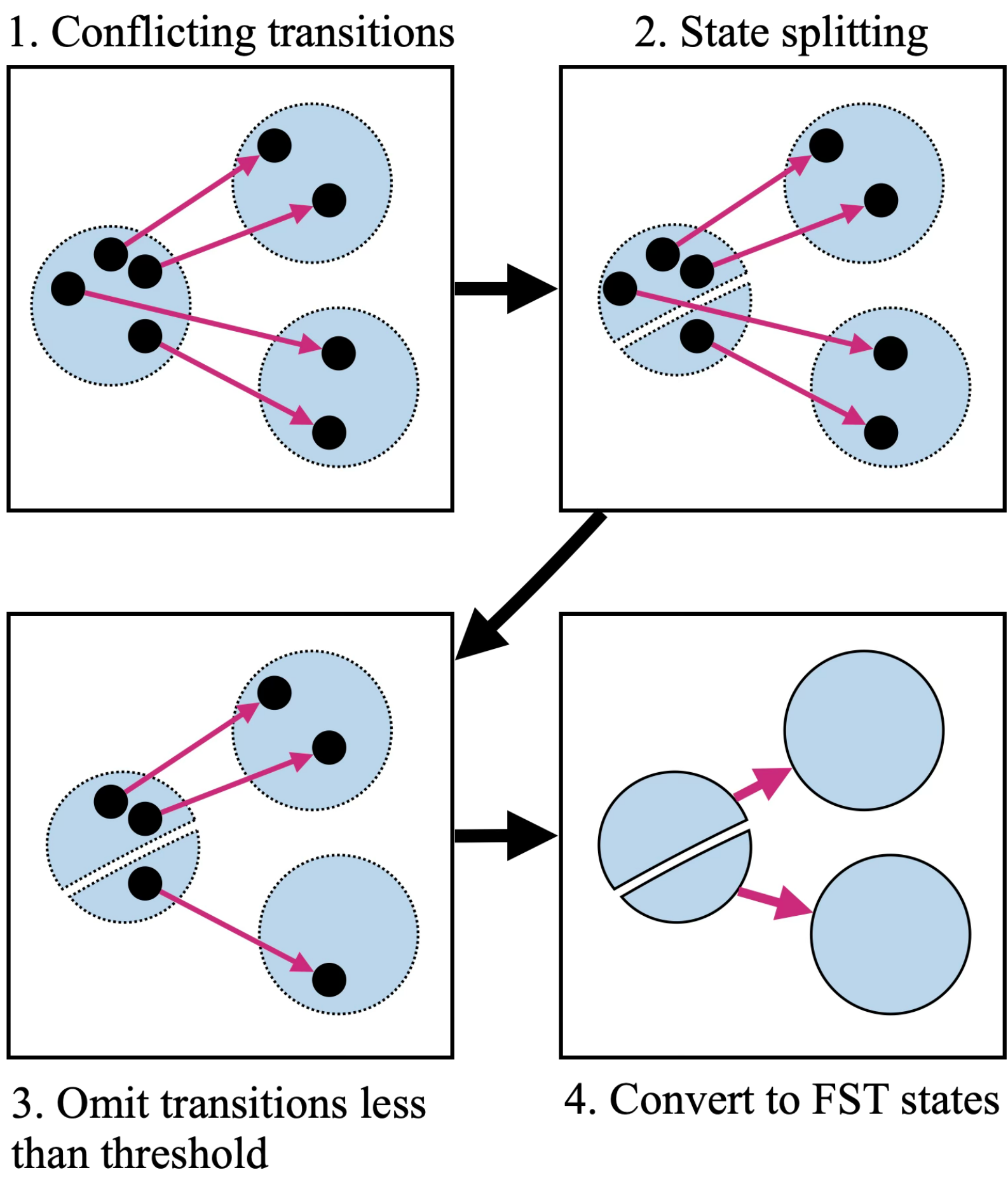}
    \caption{State splitting algorithm, where the minimum threshold $\lambda_{trans} = 2$. If a cluster has multiple possible transitions each over the threshold (1), the points are split using an SVM or logistic regression (2). If there are still conflicting transitions, but one of them is under the threshold, it is removed (3). Finally, clusters are materialized into FST states (4).}
    \label{fig:splitting}
\end{figure}

Our state splitting algorithm (diagram in \autoref{fig:splitting}, pseudocode in \autoref{sec:splitting_pseudocode}) is closely inspired by the refinement approach of \citet{weiss_extracting_2018}. Starting at the initial state, we check whether the aggregated transitions violate input determinism. Specifically, we look for multiple possible transitions for the same input symbol that occur more than a minimum threshold, $\lambda_{trans}$, a tuned hyperparameter. If there is no violation, we continue to downstream states. If there is a violation, we identify the conflicting activations and fit a classifier (either SVM or logistic regression; we try both during the hyperparameter sweep) to separate the cluster into two or more new states (including the non-conflicting activations). Since upstream states may now be input non-deterministic, we requeue these states.  After splitting concludes, we remove inaccessible states and minimize the transducer using FSA minimization that treats input/output labels as single symbols \citep{hopcroft1971n}.

\section{Experiments}
\subsection{Datasets}
We evaluate our method on benchmarks for three tasks that can be represented as regular transductions. We evaluate the constructed transducers based on exact-match accuracy for output strings. We evaluate on held-out test datasets, seeking transducers that generalize to unseen examples.\footnote{In other grammatical inference setups, the goal is to learn a minimal transducer that perfectly models the training dataset. We can trivially construct a prefix tree transducer with perfect accuracy on seen data, but that performs no generalization.}

\paragraph{Morphological inflection}
We use the 2020 SIGMORPHON Shared Task datasets for morphological inflection \citep{vylomova-etal-2020-sigmorphon}. The shared task included datasets for 90 languages;
we use the 24 languages (full breakdown in \autoref{tab:inflection_data}) used in \citet{beemer-etal-2020-linguist}, allowing us to compare to their hand-crafted transducers. Morphological inflection is formulated as predicting an inflected word-form given a lemma and morphological tags, such as the following English verb:
\begin{center}
    \texttt{[3P][Sg][Prs] run  $\rightarrow$ runs} 
\end{center}
\paragraph{Grapheme-to-phoneme}
We also use the 2020 SIGMORPHON Shared Task data on grapheme-to-phoneme prediction \citep{gorman-etal-2020-sigmorphon} for 15 languages (\autoref{tab:g2p_data}). The task consists of predicting the phonemic (using the International Phonetic Alphabet) string for a word written in the language's orthography, as in the following French example:
\begin{center}
    \texttt{hébergement $\rightarrow$ \textipa{ebE4ZEm\~A}} 
\end{center}
\paragraph{Historical normalization}
We use datasets from \citet{bollmann-2019-large} for historical normalization (which involves mapping spelling variants of the same word to a single canonical form) for seven languages (\autoref{tab:histnorm_data}). An example in historical English is given below:
\begin{center}
    \texttt{thaire $\rightarrow$ their} 
\end{center}
\subsection{Baselines}
We compare against the main algorithms for domain-agnostic transducer induction from unaligned data. 
We do not compare to domain-specific methods requiring
expert task knowledge.

\paragraph{OSTIA} OSTIA \cite{oncina_learning_1993} is a seminal algorithm for transducer induction from unaligned data.
OSTIA involves building a prefix tree for the labeled examples, pushing output substrings as early as possible, and merging states when it does not violate the input determinism condition. Transducers produced by OSTIA perfectly model the training dataset, but often struggle to generalize to other examples. Furthermore, the algorithm has polynomial 
complexity with the number of examples. We provide a highly-optimized Cython implementation,\footnote{\url{https://github.com/mhulden/pyfoma}} but even still the algorithm may take nearly 100 days to run on large datasets. For practicality, we set a time limit of one day and stop merging after reaching the limit;
these runs are marked with an asterisk (*) in the results table.

\paragraph{Data-Driven OSTIA} DD-OSTIA is a modification to OSTIA which can converge with fewer examples, by considering states 
in breadth-first order \citep{oncina1998data}. We use the variation described in \citet{higuera2010grammar}, which omits the costly equivalence measure and performs merges greedily.

\paragraph{No Change} For historical normalization, many examples do not have any changes. We compute a simple baseline that outputs the input unchanged.

\paragraph{Human Expert} For inflection, we compare against the transducers from \citet{beemer-etal-2020-linguist}, which were handcrafted by linguistics students after studying inflection datasets. These transducers are highly accurate thanks to the students' knowledge of common morphological processes, but required roughly forty hours for an expert to create. 

\subsection{Hyperparameter Tuning}
Since our system cannot be optimized end-to-end, any error from the intermediate model is propagated to the final transducer. Instead, we perform extensive hyperparameter sweeps to minimize error for each component, using a held-out evaluation set (distinct from the test set) to select the best model. Because the models involved are very small, this can be completed rapidly, with the largest languages requiring roughly a day using an NVIDIA A100 GPU. We describe the tuned hyperparameters for the RNN (\autoref{sec:rnn_training_appendix}) and extraction algorithm (\autoref{sec:extraction_appendix}).

\section{Results}
We present our main experimental comparison for inflection in \autoref{tab:inflection}, G2P in \autoref{tab:g2p}, and historical normalization in \autoref{tab:histnorm}.

\begin{table}[htb]
\centering
\begin{adjustbox}{width=0.8\linewidth}
\begin{tabular}{l | c | c c c }
\toprule 
 & Expert & OSTIA & DD-OSTIA &  Ours \\
\midrule
aka & 1.000 & 0.533 & 0.615 & \textbf{0.975} \\
ceb & 0.865 & 0.135 & 0.108 & \textbf{0.865} \\
crh & 0.964 & 0.020 & 0.005* & \textbf{0.888} \\
czn & 0.725 & 0.013 & 0.036 & \textbf{0.666} \\
dje & 1.000 & 0.125 & 0.438 & \textbf{0.750} \\
gaa & 1.000 & 0.391 & 0.331 & \textbf{0.959} \\
izh & 0.929 & 0.009 & 0.000 & \textbf{0.438} \\
kon & 0.987 & 0.141 & 0.179 & \textbf{0.846} \\
lin & 1.000 & 0.043 & 0.065 & \textbf{0.978} \\
mao & 0.667 & 0.000 & 0.048 & \textbf{0.452} \\
mlg & 1.000 & 0.189 & 0.307 & \textbf{0.969} \\
nya & 1.000 & 0.618 & 0.533 & \textbf{0.993} \\
ood & 0.710 & 0.010 & 0.013 & \textbf{0.570} \\
orm & 0.990 & 0.037 & 0.012 & \textbf{0.798} \\
sot & 1.000 & 0.424 & 0.424 & \textbf{0.727} \\
swa & 1.000 & \textbf{0.569} & 0.541 & 0.526 \\
syc & 0.883 & 0.031 & 0.007 & \textbf{0.870} \\
tgk & 0.938 & 0.000 & 0.000 & \textbf{0.938} \\
tgl & 0.778 & 0.019 & 0.013 & \textbf{0.383} \\
xty & 0.817 & 0.122 & 0.000 & \textbf{0.808} \\
zpv & 0.789 & 0.175 & 0.145 & \textbf{0.706} \\
zul & 0.833 & 0.000 & 0.000 & \textbf{0.692} \\
\bottomrule
\end{tabular}
\end{adjustbox}
\caption{Accuracy of learned transducers for morphological inflection datasets on held-out test sets. * marks OSTIA runs that reached the time limit before ending.}
\label{tab:inflection}
\end{table}
\begin{table}[htb]
\centering
\begin{adjustbox}{width=0.7\linewidth}
\begin{tabular}{l | c c c }
\toprule
& OSTIA & DD\mbox{-}OSTIA &  Ours \\
\midrule 
fre & 0.047 & 0.098 & \textbf{0.200} \\
dut & 0.009 &  0.011* & \textbf{0.149} \\
arm & 0.067 & 0.091 & \textbf{0.589} \\
geo & 0.013 & 0.056* & \textbf{0.596} \\
bul & 0.038 & 0.049 & \textbf{0.236} \\
gre & 0.087 & 0.080 & \textbf{0.307} \\
ady & 0.013 & 0.049 & \textbf{0.184} \\
kor & 0.029 & 0.033 & \textbf{0.057} \\
ice & 0.029 & 0.062* & \textbf{0.278} \\
hin & 0.029 & 0.100 & \textbf{0.411} \\
lit & 0.144 & 0.167 & \textbf{0.171} \\
jpn & 0.029 & 0.071 & \textbf{0.369} \\
rum & 0.298 & 0.271 & \textbf{0.598} \\
vie & 0.022 & 0.038 & \textbf{0.233} \\
hun & 0.053 & 0.140 & \textbf{0.522} \\
\bottomrule\end{tabular}
\end{adjustbox}
\caption{Accuracy of learned transducers for grapheme-to-phoneme datasets on held-out test sets. * marks OSTIA runs that reached the time limit.}
\label{tab:g2p}
\end{table}
\begin{table}[htb]
\centering
\begin{adjustbox}{width=0.9\linewidth}
\begin{tabular}{l | c c c c}
\toprule &  No~Change & OSTIA &   DD\mbox{-}OSTIA & Ours \\
\midrule
deu & 0.087 & \textbf{0.251*} & 0.223* & 0.214 \\
hun & 0.055 & 0.160 & 0.147* & \textbf{0.316} \\
swe & 0.431 & 0.348 & 0.338* & \textbf{0.579} \\
por & 0.395 & \textbf{0.559}  & 0.546* & 0.503 \\
slv & 0.754 & 0.644 & 0.628* & \textbf{0.801} \\
isl & 0.338 & \textbf{0.581} & 0.563* & 0.507 \\
spa & 0.550 & 0.568  & 0.566 & \textbf{0.646} \\
\bottomrule
\end{tabular}
\end{adjustbox}
\caption{Accuracy of learned transducers for historical normalization datasets on held-out test sets. * marks OSTIA runs that reached the time limit.}
\label{tab:histnorm}
\end{table}

\paragraph{Performance across Tasks}
Across tasks, our method is nearly always superior to OSTIA-based methods and simple no-change baselines. However, our method is typically more effective for inflection than the other two tasks. One possible explanation is that our RNNs are unidirectional. For inflection, the RNN can separate states based on the initial feature tags, avoiding conflicting transitions. Meanwhile, G2P often has transitions dependent on the right-side context, which the RNN is unable to distinguish. For example:

\begin{center}
    \texttt{cat} $\rightarrow$ \textipa{k\ae \textglotstop} \\
     \texttt{cent} $\rightarrow$ \textipa{s \textepsilon n t}
\end{center}

The choice of $(c,k)$ or $(c,s)$ is impossible to predict from only the left context. There are two clear paths to address this. First, an alternate alignment algorithm could avoid this issue by delaying outputs until they can be distinguished--in the prior example, the symbols could be aligned as:
\begin{center}
    \texttt{(c,$\epsilon$) (a,k\ae) (t,\textglotstop)}  \\
    \texttt{(c,$\epsilon$) (e,s\textepsilon) (n,n) (t,t)} 
\end{center}
\noindent The other option is to replace the unidirectional RNN with a bidirectional RNN (or transformer). Then, one could extract transducers by first learning a \textit{bimachine} \citep{doi:10.1137/0220042}, which can be thought of as a transducer containing two components and which emits its output depending on the joint states of both left-to-right 
and right-to-left reading components. A bimachine can then be converted to a regular FST.

Another issue arises when the input and output string have the same length, but should not be aligned with a one-to-one mapping. This occurs when there is an equal number of insertions and deletions, such as in the following G2P example ("rite"):
\begin{center}
    \texttt{(r,\textturnr) (i,\textscripta}) (t,\textsci) (e,\textglotstop)
\end{center}
\noindent Here, the preferred alignment would be:
\begin{center}
    \texttt{(r,\textturnr) (i,\textscripta \textsci}) (t,\textglotstop) (e,\textepsilon)
\end{center}
\noindent but our alignment algorithm is unable to produce this.

\paragraph{Comparison to OSTIA} While OSTIA attempts to learn generalizable transducers via greedy state merging, it does this by simply checking whether merging two states would violate an example in the training dataset. This means that two dissimilar states could be incorrectly merged so long as there is no counter-example. Meanwhile, our method leverages the hidden-state space geometry learned by the RNN, which should naturally organize so that states with similar outgoing transitions and prefixes are close. In addition, OSTIA learns to align input and output symbols by greedily pushing output symbols as early as possible in the prefix tree. However, this can result in poor generalization. Our method achieves better generalization by using
a statistical algorithm that optimizes alignments globally across the dataset.

An alternative method could use a standard neural inflection model to predict output strings for the full domain, and then use OSTIA on the expanded dataset. This 
is a coarser version of our approach, treating the neural network as a black box and disregarding any information from its hidden state geometry. Accuracy would match that 
of the best possible neural model (since OSTIA retains perfect recall on its training data), but OSTIA's reliance on naive state merging would likely produce very large and/or uninterpretable transducers.


\paragraph{Comparison to Human Expert}
In many cases, the automatically extracted transducers are competitive to human expert-created ones (and achieve perfect performance in some cases). For practical use, an expert could then correct erroneous outputs with small edits, 
saving significant effort  compared to building the transducer manually. The constructed FSTs tend to be far larger than expert-crafted transducers; for example, the inflection \texttt{ceb} constructed FST has 280 states while the expert-crafted FST has 30 states (see \autoref{sec:example_appendix}).

\section{Analysis}
We randomly selected four datasets to perform ablations: the \texttt{czn} and \texttt{kon} inflection datasets, the \texttt{geo} G2P dataset, and the \texttt{swe} normalization dataset. We use the optimal RNN hyperparameters (except for the third ablation, where we run a sweep), and we run a full sweep for FST extraction.

\subsection{Ablation: CRPAlign}
\begin{figure}[h]
    \centering
    \includegraphics[width=\linewidth]{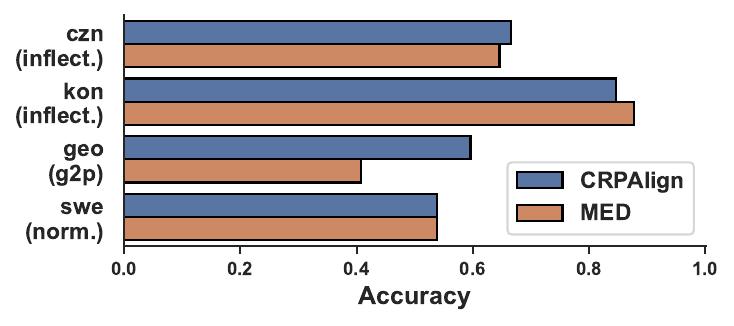}
    \caption{Scores with CRPAlign algorithm and simple minimum-edit distance (MED).}
    \label{fig:ablation1}
\end{figure}

The CRPAlign algorithm makes use of global probabilities to predict alignments. We ablate this alignment algorithm and use a simple local minimum-edit distance (MED) for a small subset of datasets in \autoref{fig:ablation1}. For the datasets where the input and output alphabets are overlapping (inflection and normalization), MED achieves very similar scores to CRPAlign. However, for G2P where the output alphabet is disjoint from the input alphabet, CRPAlign is far more effective, as MED will always need to replace the entire string.

\subsection{Ablation: Extraction with Synthetic Data}
We omit the synthetic data described in \ref{sec:hidden_states_step} and only collect activations for the training examples. For the inflection and G2P datasets, the ablated performance is worse by up to 8 points. This supports our hypothesis that collecting additional activations to attain better coverage of the input domain is beneficial to generalization. 

\begin{figure}[h]
    \centering
    \includegraphics[width=\linewidth]{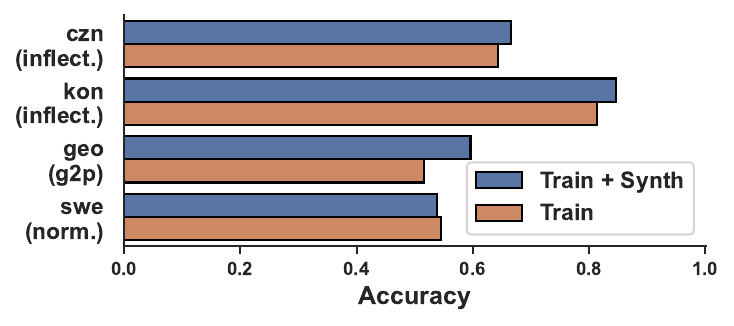}
    \caption{Scores with and without synthetic data.}
    \label{fig:ablation2}
\end{figure}

\subsection{Ablation: Transduction Objective}
\begin{figure}[h]
    \centering
    \includegraphics[width=\linewidth]{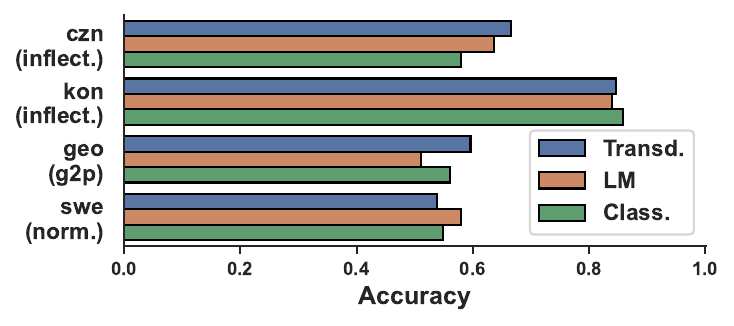}
    \caption{Scores for transducers extracted from RNNs trained for transduction, language modeling, and binary classification.}
    \label{fig:ablation3}
\end{figure}

We initially considered three different training objectives for the RNN: binary classification, next-aligned-pair prediction, and transduction (which we selected). For binary classification, we generated negative samples by randomly replacing output characters in training examples, and trained the RNN to distinguish correct and incorrect samples. For next-pair prediction, we trained a language modeling head using sequences of aligned pairs of input and output symbols. We provide results for some datasets in \autoref{fig:ablation3}, with no clear trends. While we had hypothesized that transduction would be necessary for well-separated states, these results would indicate that the process was robust to training objective.

\section{Related Work}
\label{sec:related_work}
This work is closely inspired by early work on RNNs that identified similarities to finite-state automata, providing empirical and theoretical results that suggested that RNNs trained on regular languages would learn to replicate finite-state dynamics \citep{cleeremans1989finite, giles_higher_1989, pollack_induction_1991, casey_dynamics_1996}. The basic procedure for state clustering was first proposed by \citet{giles_extracting_1991, giles_learning_1992}, though we have made heavy modifications to adapt it to transducers. Various works have proposed alternate clustering techniques such as k-means \citep{frasconi_representation_1996, gori_inductive_1998} and hierarchical clustering \citep{carbonell_hybrid_1994}. \citet{tino_learning_1995} is one of the few works that examined transducer inference, though they only studied small synthetic languages and assumed a provided alignment between inputs and outputs. \citet{schellhammer_knowledge_1998} represents the only (to the best of our knowledge) study that extracted automata from naturally-occurring datasets; here, sequences of part-of-speech tags.

More recent work has studied gated RNNs such as LSTMs, finding that it is far more difficult to extract accurate automata \citep{wang_comparative_2018,weiss_practical_2018, hou_learning_2020}. \citet{weiss_extracting_2018} proposed an effective method for active learning that repeatedly refines the states of the automaton based on a validation set, inspiring our splitting procedure. \citet{hong_adaax_2022} proposed a similar refinement algorithm that performs merging instead of splitting. Other work has attempted to infer automata from transformers with mixed success \citep{adriaensen_extracting_2024, zhang_automata_2024}, but it is unclear whether transformers are even capable of learning regular languages robustly \citep{10.1162/tacl_a_00306, bhattamishra-etal-2020-ability, liu_transformers_2023}, though hard attention transformers can be constructed to act as transducers \citep{strobl-etal-2025-transformers}.

Other work has studied weighted transducer inference using neural networks on real-world tasks such as speech recognition \citep{lecorve2012conversion} and morphological inflection \citep{rastogi-etal-2016-weighting}. Our approach could be adapted to produce weighted transducers by creating probability distributions for transitions.

\section{Conclusion}
We proposed a method for automatically constructing finite-state transducers from unaligned data, using RNNs as intermediate representations that enable transducer learning via gradient descent. We test our system on morphological inflection, grapheme-to-phoneme prediction, and historical normalization. Our system produces transducers that generalize well to unseen data on two of three tasks, far outperforming existing algorithms and approaching the accuracy of transducers created by human experts. These results provide a tractable way to create highly-performant symbolic systems, and they validate the theoretical claim that recurrent neural networks organize their hidden states with dynamics similar to finite-state automata.

\section*{Limitations}
Our method is intended for data which can be efficiently represented as an unweighted finite-state transducer. For datasets which violate Markov assumptions or require probabilistic outputs, this method will not work without modification. As noted, our method struggles on right-side dependencies. We run a wide variety of experiments on several tasks using appropriate baselines, but we do not test methods such as LLM in-context learning or reinforcement learning approaches to transducer inference. We did not have the budget to run full ablations across every dataset, which may reveal other trends.

\section*{Ethical Considerations}
Our method is intended to enable transducer inference for many languages and tasks, enabling highly performant language technology for underresourced languages. However, when using data from minority languages, there is always risk of misuse. We strive to use data as intended and do not claim ownership over this data. Furthermore, our work used significant computing resources, which carry an environmental cost.

\section*{Acknowledgments}
Foundation under Grant No. 2149404, "CAREER: From One Language to Another." Any opinions, findings, and conclusions or recommendations expressed in this material are those of the authors and do not necessarily reflect the views of the National
Science Foundation.

\bibliography{anthology-1,anthology-2,custom}

\appendix

\section{Chinese Restaurant Process Alignment}
\label{sec:alignment}

We model alignment between two symbol sequences as a \emph{monotone 1--1} path through an alignment lattice, where each step aligns a symbol pair $(a,b)$ and either side may be $\epsilon$ (insertion/deletion). Symbol-pair types that are frequent in the corpus are preferred: during training we maintain corpus-wide counts of all observed $(a,b)$ pairs and convert these counts into smoothed \emph{CRP-style} (Pólya-urn) probabilities for scoring alignments.

Rather than searching directly for a single globally optimal set of alignments, we draw samples from the posterior over alignments using a Gibbs sampler that updates one word pair at a time:

\begin{enumerate}
    \item \textbf{Initialize.} Assign each training pair an initial monotone alignment and accumulate the resulting pair-type counts over the whole corpus.
    
    \item \textbf{Gibbs sampling.} For a fixed number of iterations, sweep through the training pairs. For each pair:
    \begin{enumerate}
        \item \textbf{Remove:} subtract the counts contributed by the pair's current alignment from the global count table.
        \item \textbf{Resample:} sample a new monotone alignment from its conditional distribution given the current counts. This is implemented with dynamic programming over the lattice: a forward pass computes the total probability mass of all paths (in log space, using log-sum-exp), and a backward pass samples a single path by repeatedly choosing among diagonal (match), horizontal (insertion), and vertical (deletion) moves proportional to their path weights.
        \item \textbf{Add:} add the counts from the newly sampled alignment back into the global table.
    \end{enumerate}
    
    \item \textbf{Final alignment.} After burn-in, we average the count tables across sampled states to obtain more stable pair probabilities, and then produce a final deterministic alignment for each word pair by Viterbi decoding (minimum-cost alignment) under these averaged probabilities.\footnote{The implementation is available as a stand-alone aligner at \url{https://github.com/mhulden/crpalign}.}
\end{enumerate}

\section{RNN Training}
\label{sec:rnn_training_appendix}
We train Elman RNNs with a single layer with the following hyperparameters, where parameters with multiple values were tuned via a hyperparameter sweep. 

\begin{table}[h]
    \small
    \centering
    \begin{tabularx}{\linewidth}{l|X}
        \toprule
        Parameter & Value(s)  \\
        \midrule
        Optimizer & AdamW (default params) \\
        Activation & tanh \\
        Spec-norm weight & 0.1 \\
        Model dim. & $\{16, 32, 64, 128\}$ \\
        Dropout & $\{0, 0.1, 0.3\}$ \\
        Learning rate & $\{2E-4, 1E-3,$ $2E-3, 1E-2\}$ \\
        Label smoothing & 0.1 \\
        Batch size & $\{2\times10^k:1\leq k\leq 12\}$ \\
        Epochs & $\{200, 600, 1000\}$ \\
        \bottomrule
    \end{tabularx}
    \caption{Hyperparameters for RNN training.}
    \label{tab:rnn_hparams}
\end{table}

Batch size is limited to the four largest values that are less than one-fifth of the size of the training set. No early stopping is used, since this can result in stopping before the target metric has reached its peak \citep{choi-etal-2022-early}. The hyperparameter sweep uses Bayesian Optimization, which optimizes a surrogate model to predict which combination of hyperparameters should be selected \citep{NIPS2012_05311655}. We optimize for the validation set loss and use Hyperband with min and max epochs 75 and 1000 respectively, which allows for bad-performing runs to be stopped early \citep{JMLR:v18:16-558, kann-etal-2019-towards}. We limit to 50 runs for each dataset, except for datasets which have more than 5,000 examples, for which we limit to 25 runs. 




\section{FST Extraction Sweep}
\label{sec:extraction_appendix}
We run a Bayesian sweep over the hyperparameters for FST extraction with 100 runs for most datasets and 25 runs for datasets larger than 5,000 examples. We optimize the following parameters to maximize F1 on the evaluation set:

\begin{table}[h]
    \small
    \centering
    \begin{tabularx}{\linewidth}{l|X}
        Parameter & Value(s)  \\
        \hline
        Clustering algorithm & k-means \\
        Num. clusters & $\{n : 50 \leq n \leq n_{max}\}$ \\
        Splitting Classifier & $\{\textnormal{SVM}, \textnormal{Logistic Regression}\}$ \\
        $\lambda_{trans}$ (Split Threshold ) & $\{$None, 2, 3, 4, 5, 10, 15, 20, 25, 30, 40, 50$\}$ \\
    \end{tabularx}
    \caption{Hyperparameters for FST extraction algorithm. $n_{max}$ is the maximum number of unique hidden-state values for a given dataset. If the split threshold is None, then the splitting algorithm is not used.}
    \label{tab:extraction_hparams}
\end{table}

\section{Ablation Results}
\label{sec:ablation_appendix}
Results for the ablations are given in \autoref{tab:ablations}.

\begin{table}[h]
    \small
    \centering
    \begin{tabular}{l|c c c c}
        \toprule
        Setting & czn & kon & geo & swe  \\
        \midrule
        Base & 0.666 & 0.846 & 0.596 & 0.539 \\
        MED (Abl 1) & 0.646 & 0.878 & 0.407 & 0.539 \\
        Train only (Abl 2) & 0.643 & 0.814 & 0.516 & 0.546 \\
        LM (Abl 3) & 0.636 & 0.840 & 0.511 & 0.579 \\
        Class. (Abl 3) & 0.580 & 0.859 & 0.560 & 0.549 \\
        \bottomrule
    \end{tabular}
    \caption{Numerical results for ablation figures.}
    \label{tab:ablations}
\end{table}

\section{Datasets}
\begin{table}[tbh!]
\small
    \centering
    \begin{tabular}{c l | c c c}
        \toprule
        Code & Language & \# Train & \# Dev & \# Test \\
        \midrule
        aka & Akan  & 2,793 & 380 & 763 \\
        ceb & Cebuano  & 420 & 58 & 111 \\
        crh & Crimean Tatar  & 5,215 & 745 & 1,490 \\
        czn & Zapotecan  & 1,088 & 154 & 305 \\
        dje & Zarma  & 56 & 9 & 16 \\
        gaa & Ga & 607 & 79 & 169 \\
        izh & Ingrian  & 763 & 112 & 224 \\
        kon & Kongo  & 568 & 76 & 156 \\
        lin & Lingala  & 159 & 23 & 46 \\
        mao & Maori  & 145 & 21 & 42 \\
        mlg & Malagasy  & 447 & 62 & 127 \\
        nya & Chewa  & 3,031 & 429 & 853 \\
        ood & O'odham  & 1,123 & 160 & 314 \\
        orm & Oromo  & 1,424 & 203 & 405 \\
        ote & Mezquital Otomi  & 22,962 & 3,231 & 6,437 \\
        san & Sanskrit  & 22,968 & 3,188 & 6,272 \\
        sot & Sotho  & 345 & 50 & 99 \\
        swa & Swahili  & 3,374 & 469 & 910 \\
        syc & Syriac  & 1,917 & 275 & 548 \\
        tgk & Tajik  & 53 & 8 & 16 \\
        tgl & Tagalog  & 1,870 & 236 & 478 \\
        xty & Yoloxóchitl Mixtec  & 2,110 & 299 & 600 \\
        zpv & Chichicapan Zapotec  & 805 & 113 & 228 \\
        zul & Zulu  & 322 & 42 & 78 \\
         \bottomrule
    \end{tabular}
    \caption{Inflection datasets from \citet{vylomova-etal-2020-sigmorphon}.}
    \label{tab:inflection_data}
\end{table}

\begin{table}[tbh!]
\small
    \centering
    \begin{tabular}{c l | c c c}
        \toprule
        Code & Language & \# Train & \# Dev & \# Test \\
        \midrule
         ady & Adyghe  & 3,600 & 450 & 450 \\
        arm & Armenian   & 3,600 & 450 & 450 \\
        bul & Bulgarian  & 3,600 & 450 & 450 \\
        dut & Dutch  & 3,600 & 450 & 450 \\
        fre & French  & 3,600 & 450 & 450 \\
        geo & Georgian  & 3,600 & 450 & 450 \\
        gre & Modern Greek & 3,600 & 450 & 450 \\
        hin & Hindi  & 3,600 & 450 & 450 \\
        hun & Hungarian  & 3,600 & 450 & 450 \\
        ice & Icelandic  & 3,600 & 450 & 450 \\
        jpn & Japanese  & 3,600 & 450 & 450 \\
        kor & Korean & 3,600 & 450 & 450 \\
        lit & Lithuanian  & 3,600 & 450 & 450 \\
        rum & Romanian  & 3,600 & 450 & 450 \\
        vie & Vietnamese  & 3,600 & 450 & 450 \\
         \bottomrule
    \end{tabular}
    \caption{G2P datasets \citep{gorman-etal-2020-sigmorphon}.}
    \label{tab:g2p_data}
\end{table}

\begin{table}[tbh!]
\small
    \centering
    \begin{tabularx}{\columnwidth}{c X | c c c}
        \toprule
        Code & Language & \# Train & \# Dev & \# Test \\
        \midrule
         deu & German - Anselm & 8,268 & 8,242 & 10,000* \\
         hun & Hungarian  & 46,907 & 8,987 & 8,946 \\
         isl & Icelandic  & 10,864 & 2,490 & 2,470 \\
         por & Portuguese - 19th c. & 15,699 & 3,544 & 3,670 \\
         slv & Slovenian - Gaj & 35,436 & 8,330 & 8,614 \\
         spa & Spanish - 19th c. & 5,577 & 1,086 & 1,204 \\
         swe & Swedish & 8,465 & 1,308 & 9,576 \\
         \bottomrule
    \end{tabularx}
    \caption{Normalization datasets \citep{bollmann-2019-large}}
    \label{tab:histnorm_data}
\end{table}

We report statistics for each of the datasets used in \autoref{tab:inflection_data}, \autoref{tab:g2p_data} and \autoref{tab:histnorm_data}. For historical normalization, if there were multiple distinct sources (often based on different time periods), we selected on source.

\section{Clustering and Splitting Algorithm}
The pseudocode for the splitting algorithm is given in \autoref{sec:splitting_pseudocode}.
\label{sec:splitting_pseudocode}
\begin{algorithm*}
\caption{State clustering and splitting. Assume $\textsc{Trans}(q, \sigma)$ gives the transitions for state $q$ and input symbol $\sigma$, returned as a set of tuples $(\pi, q_{dest})$ with the output symbol and destination state. Likewise, \textsc{TransOverThreshold} returns only transitions that occur more than the threshold. \textsc{SVM} splits a state to resolve multiple possible transitions for a given input.}
\label{cluster_split}
\begin{algorithmic}[1]
\Procedure{ResolveTransitions}{$q_0$, $\lambda_{trans}$}
\State $\textit{Q} \gets [q_0]$
\State $\textit{visited} \gets \{\}$

\While {length of \textit{Q} > 0}
\State $q \gets Q_0$
\State $Q \gets Q_{1...|Q|}$
\For {$\sigma \in \Sigma$}
\State $T_q \gets \textsc{TransOverThreshold}(q, \sigma, \lambda_{trans})$
\If{$|T_q| = 0$}
\State continue
\ElsIf{$|T_q| = 1$}
\State $\textsc{Trans}(q, \sigma) \gets T_q$
\State $(\pi, q_{dest}) \gets {T_q}_0$
\State $Q \gets Q \cdot [q_{dest}]$
\Else
\State $Q_{upstream} \gets \textsc{Split}(q)$
\State $Q \gets Q \cdot Q_{upstream}$
\EndIf
\EndFor

\EndWhile
\EndProcedure

\Procedure{Split}{$q$, $\lambda_{trans}$}
\State $\sigma_{worst} = \textnormal{argmax}_{\sigma} |\textsc{TransOverThreshold}(q, \sigma, \lambda_{trans})|$
\State $Q_{new} \gets \textsc{SVM}(q, \sigma_{worst})$
\State return $ \{q_{source} : [\exists q_{new} \in Q_{new} :(*, q_{new}) \in \textsc{Trans}(q_{source}, *)]\}$
\EndProcedure
\end{algorithmic}
\end{algorithm*}

\section{Example Transducers}
\label{sec:example_appendix}
\begin{figure*}
    \centering
    \includegraphics[width=\linewidth]{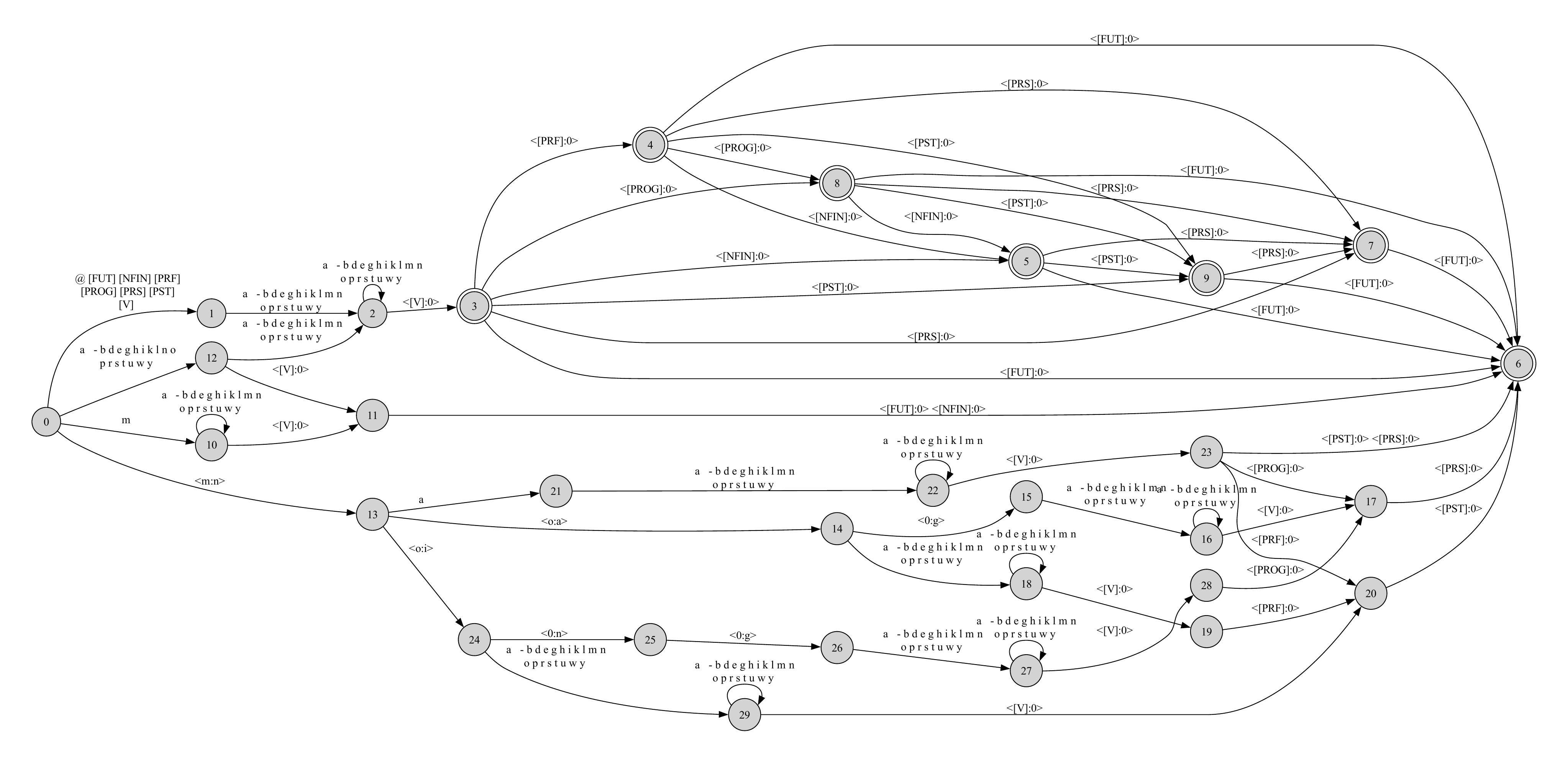}
    \caption{Expert-crafted FST for inflection \texttt{ceb} dataset.}
    \label{fig:expert_ceb}
\end{figure*}

\begin{figure*}
    \centering
    \includegraphics[width=\linewidth]{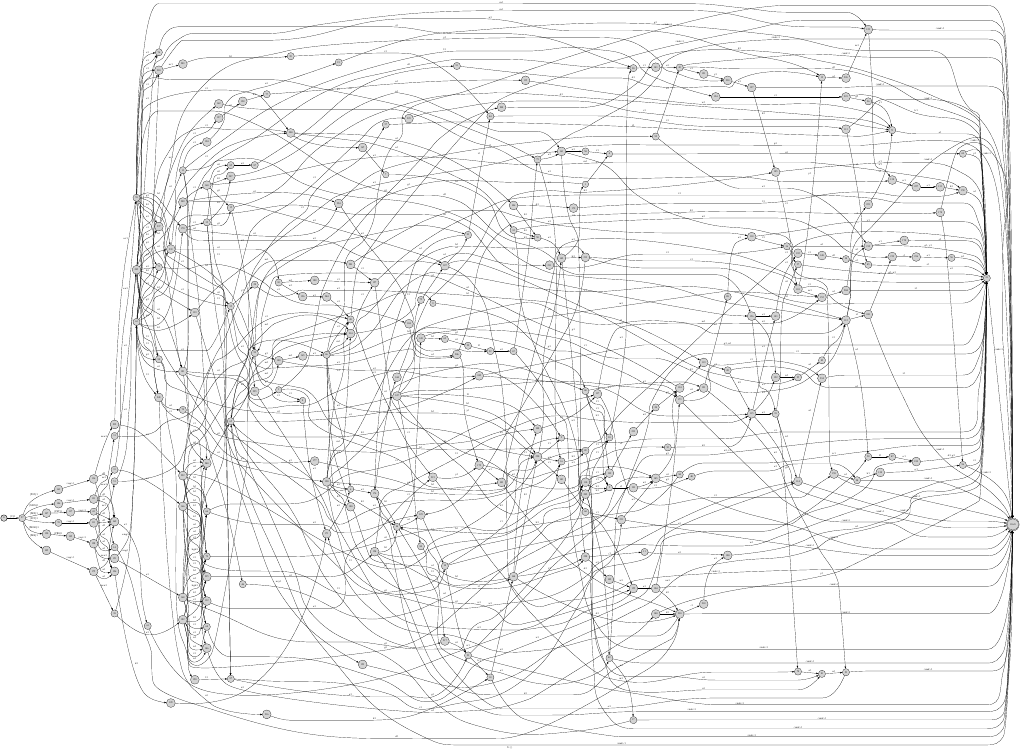}
    \caption{Automatically-constructed FST for inflection \texttt{ceb} dataset.}
    \label{fig:auto_ceb}
\end{figure*}

We visualize transducers for the inflection \texttt{ceb} dataset, comparing the expert-crafted (\autoref{fig:expert_ceb}) and automatically constructed (\autoref{fig:auto_ceb}) transducers.

\clearpage

\end{document}